\documentclass[10pt,twocolumn,letterpaper]{article}

\usepackage[pagenumbers]{iccv} 

\usepackage[numbers,sort&compress]{natbib}

\definecolor{iccvblue}{rgb}{0.21,0.49,0.74}
\usepackage[pagebackref,breaklinks,colorlinks,allcolors=iccvblue]{hyperref}
\usepackage{comment}
\usepackage{booktabs}
\usepackage{tabularx}
\usepackage{stfloats}
\usepackage{float}

\def\paperID{7} 
\def\confName{ICCV}
\def\confYear{2025}

\title{Paired-Sampling Contrastive Framework for Joint Physical-Digital Face Attack Detection}

\author{
Andrei Balykin\thanks{Equal contribution} \\
IDRND \\
{\tt\small andrew.balykin@idrnd.net}
\and
Anvar Ganiev\footnotemark[1] \\
IDRND \\
{\tt\small anvar.ganiev@idrnd.net}
\and
Denis Kondranin\footnotemark[1] \\
IDRND \\
{\tt\small denis.kondranin@idrnd.net}
\and
Kirill Polevoda \\
IDRND \\
{\tt\small kirill.polevoda@idrnd.net}
\and
Nikolai Liudkevich \\
IDRND \\
{\tt\small lyudkevich@idrnd.net}
\and
Artem Petrov \\
IDRND \\
{\tt\small petrov@idrnd.net}
}

\begin{document}
\maketitle
\begin{abstract}
Modern face recognition systems remain vulnerable to~spoofing attempts, including both physical presentation attacks and digital forgeries. Traditionally, these two attacks vectors have been addressed by separate models or~pipelines, each targeted to its specific artifacts and modalities. However, maintaining distinct detectors leads to increased system complexity,  higher inference latency, and a combined attack vectors. We propose the Paired-Sampling Contrastive Framework, a unified training approach that leverages automatically matched pairs of genuine and attack selfies to learn modality-agnostic liveness clues. Evaluated on the 6th Face Anti-Spoofing Challenge “Unified Physical-Digital Attack Detection” benchmark, our method obtained an average classification error rate (ACER) of 2.10\%, outperforming prior solutions.  The proposed framework is lightweight, requires only 4.46 GFLOPs and a training runtime under one hour, making it practical for real-world deployment.
Code and pretrained models are available at \url{https://github.com/xPONYx/iccv2025_deepfake_challenge}.

\end{abstract}    
\section{Introduction}
\label{sec:intro}
eKYC (electronic Know Your Customer) systems have become ubiquitous in modern security and authentication applications, ranging from mobile device unlocking to financial transactions. Despite significant advances in accuracy and speed, these systems remain vulnerable to a broad spectrum of attacks. On the one hand, \emph{presentation attacks}, where an adversary presents a physical artifact such as a printed photo, replayed video, or 3D mask to the sensor, can circumvent liveness checks by exploiting superficial visual clues. On the other hand, \emph{digital attacks} leverage generative models to create highly realistic face forgeries at the pixel level, enabling identity swaps, facial expression animation, lip-syncing, and other manipulations that are difficult to spot even by expert observers. 

\begin{figure}[t]
    \centering
    \includegraphics[width=1.0\linewidth]{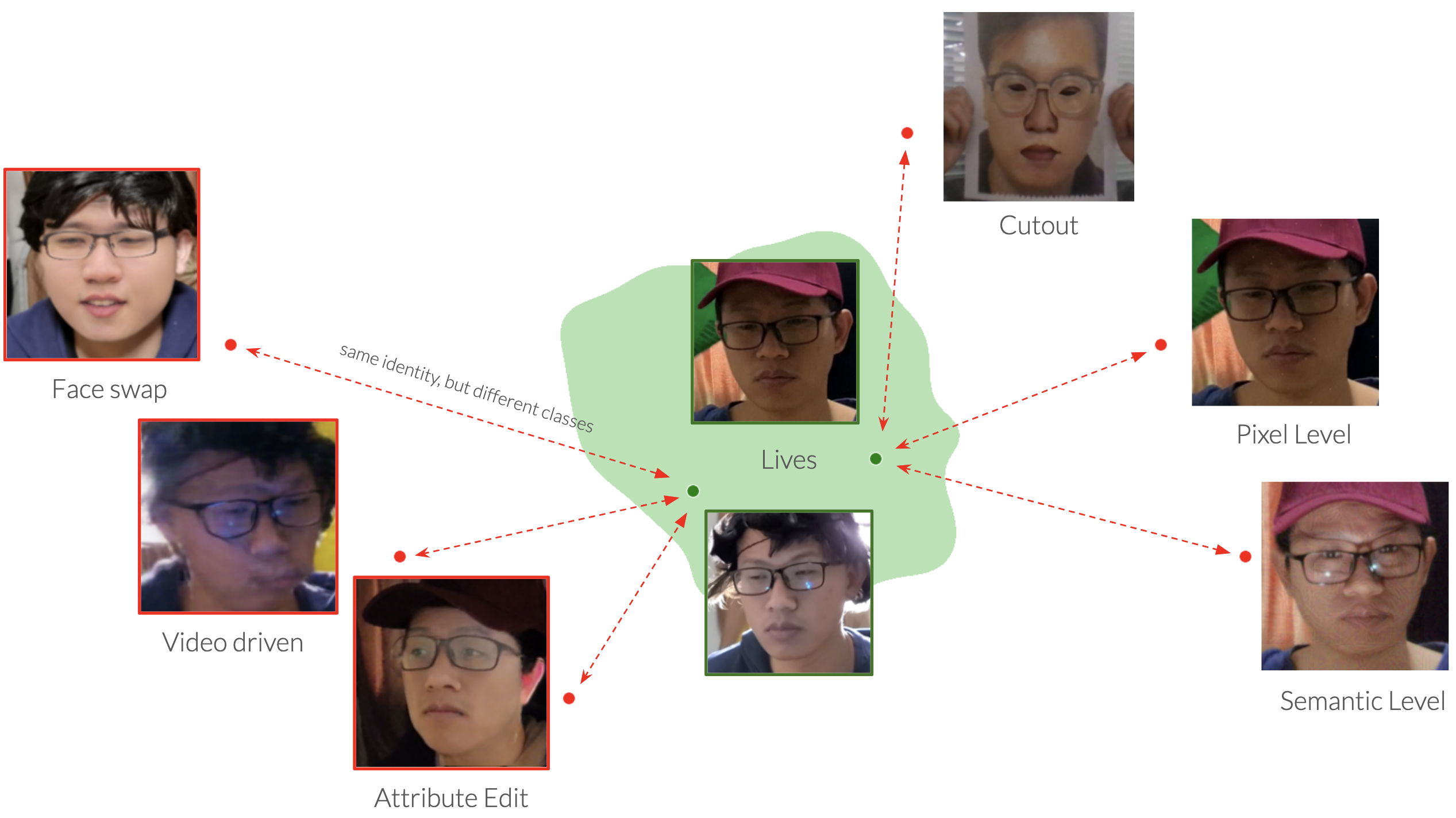}
    \caption{At the core of proposed framework is the sampling of live–attack pairs with the same user identity, which is essential to counteract bias introduced by the limited number of genuine samples in the dataset.}
    \label{fig:scheme}
\end{figure}

Due to both types of attack continuing to advance, with physical spoofs improving in quality and deepfake generators producing fewer visual artifacts, it is no longer sufficient to treat presentation attack detection and deepfake detection as separate tasks. A unified defense must accurately detect any attempt to fool a face recognition system, whether through an in-front-of-camera spoof or a digitally manipulated video. Recent benchmarks, such as the ICCV 2025 "The 6th Face Anti-Spoofing: Unified Physical-Digital Attacks Detection" track and its associated UniAttackData dataset \cite{Fang2024UniAttackData}, have underscored the importance and difficulty of this joint problem. Models trained exclusively on PAD data often fail on digital forgeries, and vice versa, due to disparate cues and domain gaps \cite{Yu2024JFSFDB}.  

In this work, we propose a \emph{Paired-Sampling Contrastive } framework that unifies PAD and DFD under a single-model architecture. Our key idea presented on Fig.~\ref{fig:scheme} is to form matched pairs of genuine and attack face embeddings, and to apply \emph{asymmetric augmentations} only to the genuine samples. A ConvNeXt-v2‐Tiny backbone then optimizes both a binary classification loss (to separate live vs.\ attack) and a supervised contrastive loss (to pull genuine features together and push all attack features apart), while CutMix regularization simulates partial occlusions and encourages the network to rely on distributed liveness cues. Challenge results demonstrate that our approach achieves an average classification error rate (ACER) of 2.10\%, ranking among the top-performing solutions.

\section{Related work}

Image-based physical presentation attacks and digital forgeries detection approaches analyze single images to identify spoofing without relying on temporal cues. Recent advancements span diverse model architectures, generalization strategies, and benchmark datasets.

\textbf{Model Architectures.} CNN-based methods, particularly EfficientNets~\cite{tan2019efficientnet}, have been popular due to their strong feature extraction capabilities and robustness to subtle visual artifacts~\cite{george2019biometric}. However, CNN models often face limitations in generalizing beyond training datasets. Vision Transformers (ViTs) \cite{dosovitskiy2020image} have recently emerged as effective alternatives, leveraging long-range dependencies within images to detect manipulation inconsistencies. For instance, Luo et al. introduced adaptive modules on pre-trained ViTs, significantly enhancing cross-dataset detection accuracy~\cite{luo2024forgery}. Multi-task models like LAA-Net combine CNN backbones with attention mechanisms, explicitly focusing on localized blending artifacts to identify high-quality deepfakes~\cite{Nguyen2024LAANet}. Additionally, hybrid frameworks such as TruFor fuse visual features with noise residuals through transformer architectures, improving detection across diverse forgery types~\cite{Guillaro2023TruFor}.

\textbf{Generalization Strategies.} A core challenge remains the generalization across different deepfake datasets. Data augmentation techniques like compression, blurring, and color adjustments effectively enhance model robustness~\cite{Dong_2023_CVPR}. In \cite{Shiohara2022SBI}, the authors demonstrated improved generalization by generating synthetic data through blending real faces, creating forgery-like artifacts without deepfake-specific models. Identity-invariant learning also plays a crucial role by reducing implicit identity bias, ensuring models focus on manipulation artifacts rather than specific facial identities~\cite{Huang2023ImplicitIdentity}.

\textbf{Datasets.} Standardized benchmarks drive the evaluation of detection methods. FaceForensics++ is widely utilized for initial training, but its limitations necessitate testing on more challenging sets like Celeb-DF and DFDC to assess generalization capabilities~\cite{Rossler2019FaceForensicsPlusPlus,Li2020CelebDF,Dolhansky2020DFDC}. Other datasets such as WildDeepfake offer real-world complexities that further challenge models to generalize effectively beyond controlled scenarios~\cite{Zi2021WildDeepfake}.

\textbf{Joint Physical and Digital Attack Datasets.} Emerging threats have expanded the scope of deepfake detection research to simultaneously address both digital manipulations and physical presentation attacks (PAD). These combined datasets integrate diverse attack vectors, providing richer benchmarks that test detectors' robustness across various attack modalities. GrandFake~\cite{Deb2023GrandFake} merges PAD and digital deepfake datasets, encompassing a broad spectrum of 25 distinct attack types. This comprehensive dataset aids in developing unified detection methods capable of identifying diverse forgery and spoofing attempts simultaneously. Similarly, in~\cite{Yu2024JFSFDB}, nine different PAD and forgery datasets were consolidated, emphasizing the critical issue of identity confounders. It facilitates the analysis of detectors' vulnerability to identity-specific biases and supports strategies aimed at creating identity-invariant detection models. UniAttackData~\cite{Fang2024UniAttackData} presents a notable advancement by maintaining identity consistency across both physical and digital attacks for each subject. This design uniquely enables models to learn consistent identity-based features, thus enhancing generalization and robustness across various types of manipulations.

In summary, image-based physical presentation attacks and digital forgeries detection methods have evolved significantly through the integration of advanced architectures and generalization strategies. Continued progress relies on combining these innovations to maintain robustness against increasingly sophisticated forgeries.
\section{Methodology}

In this section we outline the three core components of the proposed approach. Sec.~\ref{sec:data_filtering} introduces our data filtering pipeline. Subsequently, Sec.~\ref{sec:sampling_setup}   contains details of the paired setup employed during training. Finally, Sec.~\ref{sec:contrastive_training}   explains the contrastive training procedure used for representation learning.

\subsection{Data filtering}
\label{sec:data_filtering}

During manual analysis of the training dataset, we encountered a significant number of live samples that did not contain faces or any valuable information. To ensure the training setup includes only valid live samples, we manually filtered the live subset, removing approximately 10\% of non-valid images from the original set.

To eliminate potential bias from identities that appear only in the spoofing subset, we propose a filtering pipeline based on a face recognition system. First, we extract 512-dimensional face embeddings for all training images using a pretrained InceptionResNet-v1 model \cite{szegedy2017inception}. In parallel, we obtain face bounding boxes with the MTCNN detector \cite{zhang2016joint}.

Next, we compute the cosine similarity between each spoofing attack sample and all live samples available in the training data:

\[
\operatorname{sim}(a,b)
   = \frac{\langle a,\,b\rangle}{\|a\|_2\,\|b\|_2},
\]
where \(a,b \in \mathbb{R}^{512}\) are the InceptionResNet-v1 face embeddings
of a spoof-attack image and a live image, respectively.

For each spoofing attack sample we find live sample with the highest similarity:
\[
b^{\star}(a)
   = \arg\max_{\,l\in\mathcal{L}} \operatorname{sim}\bigl(a,l\bigr),
\]
with \(\mathcal{L}\) denoting all live embeddings in the training set.

To proceed with filtering, we keep a spoofing attack sample only if its similarity to the most similar live sample exceeds a predefined threshold:
\[
\mathcal{D}_{\text{train}}
   = \bigl\{ \,(a,\,b^{\star}(a)) \;\bigl|\;
        a \in \mathcal{A},\;
        \operatorname{sim}\bigl(a,\,b^{\star}(a)\bigr) > \tau_{sim}
     \bigr\},
\]
where \(\mathcal{A}\) is the set of all attack embeddings and \(\tau_{sim}\) is the
cosine-similarity threshold. We treat similarity as a training hyperparameter that can be optimized based on the target metric on the validation subset.

As a result of the filtering procedure (Tab.~\ref{tab:filtering-stats}), certain types of presentation attacks were entirely removed from the training set due to the lack of sufficiently similar live counterparts. In particular, all PAD attacks such as Replay, Print, and Cutouts were completely filtered out. Furthermore, digital attack types such as Face Swap and Attribute Edit were almost entirely eliminated, with only a small fraction of samples remaining after similarity-based filtering.

\begin{table}[ht]
\centering
\begin{tabular}{|l|r|r|}
\hline
\textbf{Category}       & \textbf{Before Filtering} & \textbf{After Filtering} \\
\hline
Pixel-Level            & 8\,364   & 5\,024  \\
Semantic-Level          & 3\,757   & 2\,349  \\
Video-Driven            & 1\,540   & 520   \\
Face-Swap               & 6\,160   & 20    \\
Attribute-Edit           & 1\,476   & 5     \\
Replay                  & 109    & --    \\
Cutouts                 & 79     & --    \\
Print                   & 43     & --    \\
\hline
Live Face               & 839    & 751   \\
\hline
\end{tabular}
\caption{Sample distribution across attack categories in train subset before and after proposed filtering.}
\label{tab:filtering-stats}
\end{table}

\subsection{Sampling setup}
\label{sec:sampling_setup}
Motivated by the large number of pixel-level, adversarial, and attribute-editing attacks that preserve the original identity while introducing subtle manipulations, we aim to improve model sensitivity to these fine-grained artifacts. To stabilize training on a highly constrained set of live samples and increase attention to distinguishing such artifacts, we leverage paired-sampling of live and attack images within each training batch.

\begin{figure}[t]
    \centering
    \includegraphics[width=1.0\linewidth]{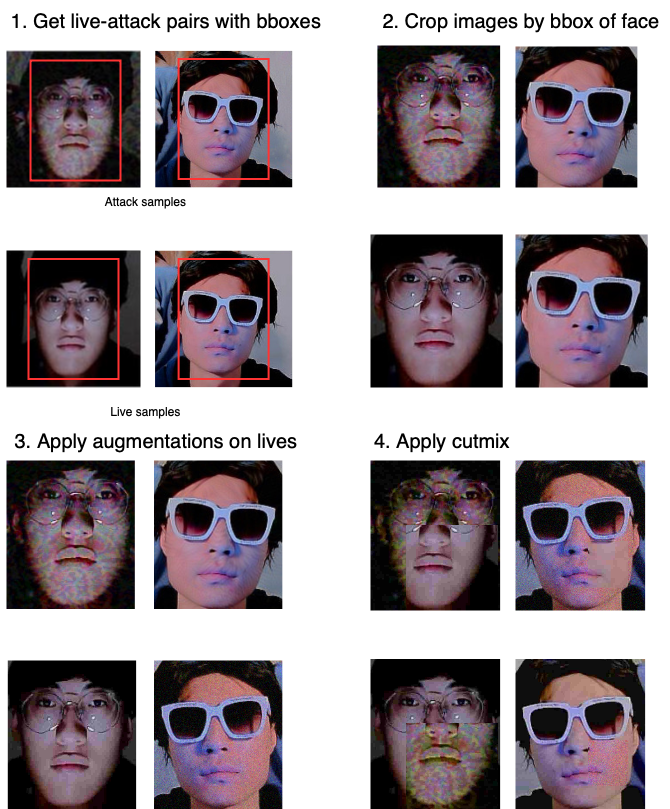}
    \caption{Overview of our data preprocessing steps used to obtain the final training pairs in our proposed pipeline: (1)\ Pairing attacks with corresponding lives. (2)\ Cropping image with bounding boxes. (3)\ Applying augmentations on live samples. (4)\ Applying CutMix augmentation strategy}
    \label{fig:my_image}
\end{figure}

We utilize the previously obtained mapping between each attack sample and its most similar live counterpart in our proposed approach. First, we sample half of the batch with spoofing attack samples, and then we append the most similar live sample for each attack in the batch. The final set of training pairs consists of 7\,918 pairs.

Let $\mathcal{S}$ denote the set of spoofing attack samples and $\mathcal{L}$ the set of live samples in the training dataset. If the cardinality of the spoofing subset is significantly greater than that of the live subset, i.e.,
\[
|\mathcal{S}| \gg |\mathcal{L}|,
\]
then the proposed pairing strategy can act as an implicit oversampling mechanism for the live subset. Since the most similar live sample is selected for each spoofing sample, live samples from the smaller cohort $\mathcal{L}$ are more likely to be sampled repeatedly, thereby increasing their influence during training.

\subsection{Contrastive training}
\label{sec:contrastive_training}
To improve the discriminative power of the learned representation we optimise a dual-objective consisting of Focal loss~\cite{lin2017focal}, and a supervised contrastive SupCon loss~\cite{khosla2020supervised}. Let $\mathbf{x}$ denote an input image, $\mathbf{f}_{\theta}(\mathbf{x})\in\mathbb{R}^{d}$ the backbone feature vector, and $\mathbf{z}=\operatorname{proj}\!\bigl(\mathbf{f}_{\theta}(\mathbf{x})\bigr)\in\mathbb{S}^{d-1}$ its $\ell_2$-normalised projection on the unit sphere. For a mini-batch $\mathcal{B}=\{(\mathbf{z}_{i},y_{i})\}_{i=1}^{N}$ the SupCon term for an anchor $i$ is:

\begin{align}
\mathcal{L}_{\text{supcon}}^{(i)}
&=
-\log
\frac{
\displaystyle \sum_{j\neq i}\mathbf{1}[y_{j}=y_{i}]\,\exp\!\left(\frac{\mathbf{z}_{i}^{\top}\mathbf{z}_{j}}{T}\right)}
{
\displaystyle \sum_{k\neq i}          \exp\!\left(\frac{\mathbf{z}_{i}^{\top}\mathbf{z}_{k}}{T}\right)}
\\[1ex]
\mathcal{L}_{\text{supcon}}
&=
\frac{1}{N}\sum_{i=1}^{N}\mathcal{L}_{\text{supcon}}^{(i)},
\end{align}

where $T$ is the temperature hyperparameter. 

\section{Experiments}

\subsection{Experimental Setting}

\textbf{Dataset and protocol.}
All experiments are conducted on the train and development pack released for The 6th Face Anti-Spoofing Challenge – Unified Physical-Digital Attack Detection@ICCV 2025 (FAS-ICCV 2025). The dataset contains \textbf{23\,367  images}, only \textbf{839} images are bona fide, while the remaining \textbf{21\,528} represent spoofed samples from nine diverse attack categories. These include both presentation attacks (print, replay, cutouts) and digital manipulations (attribute editing, face swap, video-driven synthesis, and both pixel- and semantic-level adversarial perturbations).


\textbf{Data augmentation.}
To mitigate class imbalance and improve generalization, we apply an Albumentations-based augmentation pipeline during training:
\begin{itemize}
    \item Geometric: horizontal flip,
    \item Photometric: brightness/contrast ($\pm$10\%), hue/saturation ($\pm$10), gamma (range 80–120),
    \item Compression: JPEG quality sampled from [40, 60].
\end{itemize}
Additionally, we apply CutMix augmentation with probability${=}0.3$, and $\alpha{=}0.6$ to both live and attack samples, which encourages the model to localize spoof-specific artifacts and mitigates overfitting.


\textbf{Network architecture.}
Our model backbone is \textbf{ConvNeXt-v2-Tiny} (28M parameters) \cite{convnextv2}, pretrained on  ImageNet1K \cite{deng2009imagenet}. Each input face-cropped image is resized to \textbf{$224 \times 224 \times 3$} (standard RGB), globally average-pooled by the backbone, and then passed to two heads:
\begin{enumerate}
\item a \textit{binary classification head}, single fully-connected layer for live/attack prediction
\item a \textit{projection head}, two-layer MLP that produces 128-dimensional embeddings for contrastive training.

\end{enumerate}


\textbf{Loss function.}
Training minimizes a weighted sum of focal and supervised contrastive loss:
\[
\mathcal{L} = \mathcal{L}_{\text{focal}} + \lambda\,\mathcal{L}_{\text{supcon}}
\]
The training objective combines focal loss with a class-balancing factor~$\alpha = 0.5$ and focusing parameter~$\gamma = 0.7$ to counter label imbalance and a supervised contrastive loss evaluated over the entire mini-batch of identity-consistent live–attack pairs.

\begin{figure}[t]
    \centering
    \includegraphics[width=1.0\linewidth]{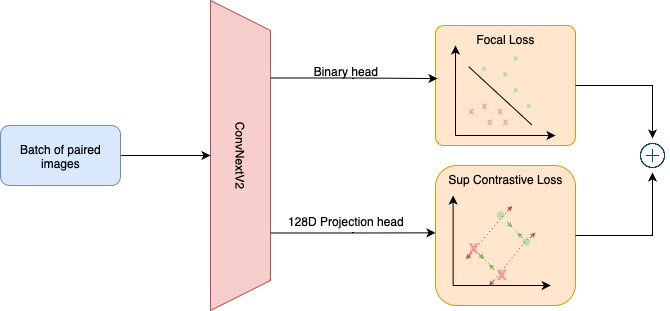}
    \caption{The proposed combination of loss functions used in our training strategy.}
    \label{fig:my_image}
\end{figure}

We use CutMix-augmented inputs and their mixed labels exclusively for the Focal loss head, while the SupCon head operates on original, unperturbed labels. This separation prevents the mixed-label targets produced by CutMix from corrupting the positive mask of the SupCon loss.


\paragraph{Evaluation metrics.}
The competition leaderboard is ranked by the \textbf{Average Classification Error Rate (ACER)} on the hidden test set.  
For completeness we additionally report
\begin{itemize}[leftmargin=*,nosep]
    \item \textbf{EER} -- the Equal-Error Rate at the ROC operating point where $\mathrm{FAR}=\mathrm{FRR}$;
    \item \textbf{APCER} -- overall Attack Presentation Classification Error Rate;
    \item \textbf{BPCER} -- overall Bona-Fide Presentation Classification Error Rate;
    \item \textbf{ACER} -- the mean of APCER and BPCER values;
\end{itemize}
The checkpoint with the lowest validation EER is selected for final submission.

\vspace{1ex}
\noindent The metrics are defined as:
\[
\text{EER:}\quad \text{FAR} = \text{FRR},
\]
\[
\text{ACER} = \frac{\text{APCER} + \text{BPCER}}{2},
\]

\[
\begin{aligned}
\mathrm{APCER} &= \frac{\mathrm{FP}}{\mathrm{FP} + \mathrm{TN}},\\[6pt]
\mathrm{BPCER} &= \frac{\mathrm{FN}}{\mathrm{FN} + \mathrm{TP}}.
\end{aligned}
\]

\paragraph{Training details.}
We fine-tuned the backbone on resized face crops. For data filtering, we used $\tau_{sim}=0.9$ as cosine similarity threshold. Mini-batches of 32 images are processed on 2x NVIDIA L4 24GB GPU; each nominal epoch spans half of the training set, and the run lasts 20 epochs.  Optimization employs \emph{AdamW} with weight-decay $1.1\times10^{-5}$.  A 5\% warm-up cosine schedule anneals the rate from $1.82\times10^{-4}$ to $6.8\times10^{-7}$ in a single cycle.  The objective combines binary focal loss ($\alpha=0.5$, $\gamma=0.7$) with supervised contrastive loss (projection dimension 128, temperature $\tau=0.14$, weight $\lambda=0.3$). CutMix applied to both live and attack crops with probability 0.3 and $\alpha=0.6$.  The checkpoint achieving the lowest validation EER was retained for submission.

\subsection{Evaluation}
We evaluated the Paired Liveness Contrastive Framework on the official validation split of "The 6th Face Anti-Spoofing: Unified Physical-Digital Attacks Detection@ICCV 2025".
Experiments were conducted with and without key components: data pairing, asymmetric live‐frame augmentation, and supervised contrastive learning,  whose individual impacts on ACER reduction are detailed in Tab.~\ref{tab:convnext-fullwidth}. Our proposed approach demonstrated 2.10\% ACER on the hidden test set of the challenge. 
Together, these results validate the effectiveness and practicality of our unified framework for simultaneous detection of both physical and digital face attacks.


\begin{table}[t]
\centering
\small
\renewcommand{\arraystretch}{1.05}
\begin{tabularx}{\columnwidth}{c c c c} 
\toprule
\textbf{Setup} & \textbf{ACER\,$\downarrow$} & \textbf{Acc\,$\uparrow$} & \textbf{AUC\,$\uparrow$} \\
\midrule
w Lives Augs \& SupCon      & 0.0085 & 0.9831 & 0.9910 \\
w/o SupCon                   & 0.0177 & 0.9646 & 0.9871 \\
w/o Lives Augmentations      & 0.0862 & 0.9046 & 0.9646 \\
\bottomrule
\end{tabularx}
\caption{Validation results under different training setups.}
\label{tab:convnext-fullwidth}
\end{table}

\subsection{Ablation Study}

\begin{figure*}[t]
  \centering
  \begin{subfigure}{0.49\textwidth}
    \centering
    \includegraphics[width=\linewidth]{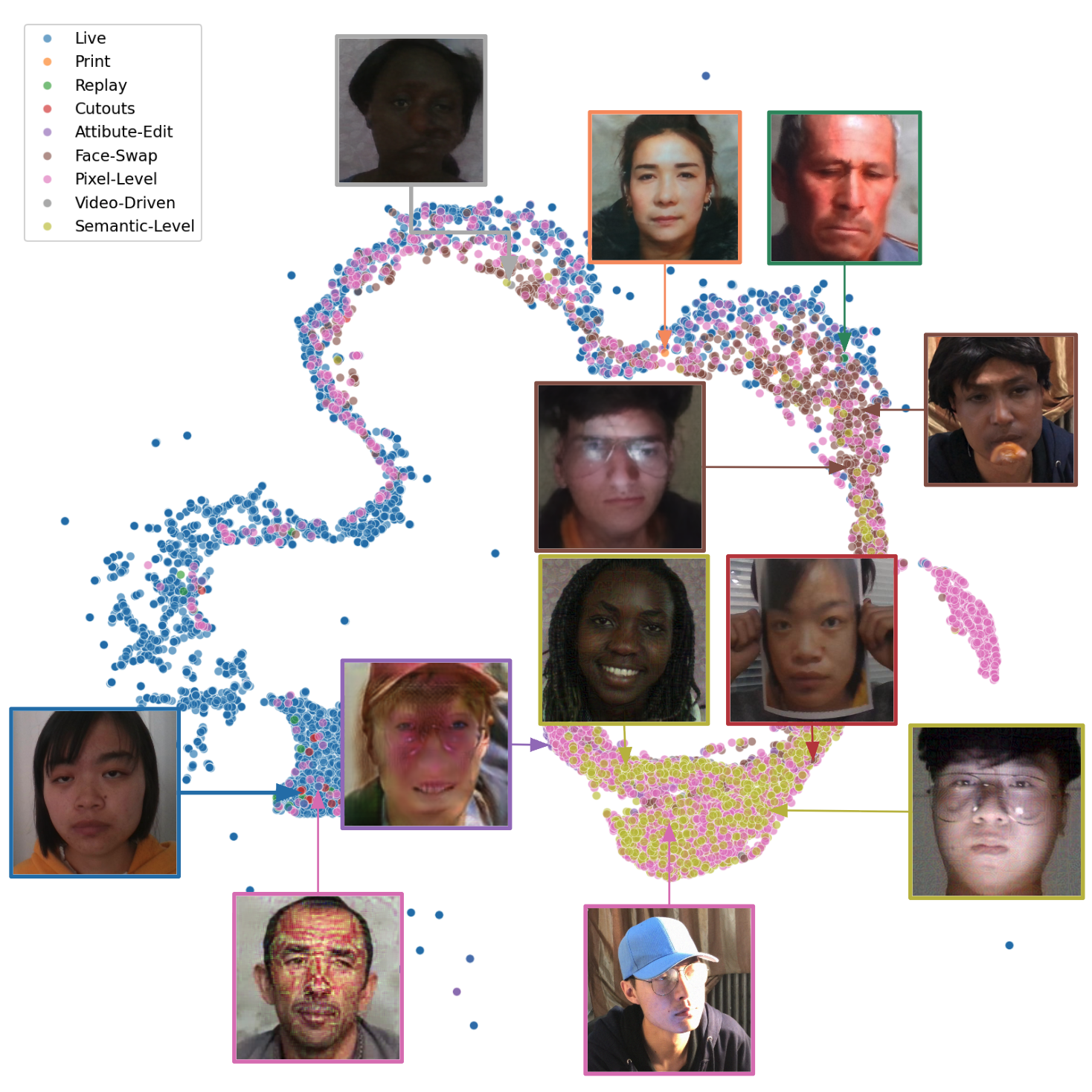}
    \caption{Learned feature space for different classes}
    \label{fig:umap-val}
  \end{subfigure}\hfill
  \begin{subfigure}{0.49\textwidth}
    \centering
    \includegraphics[width=\linewidth]{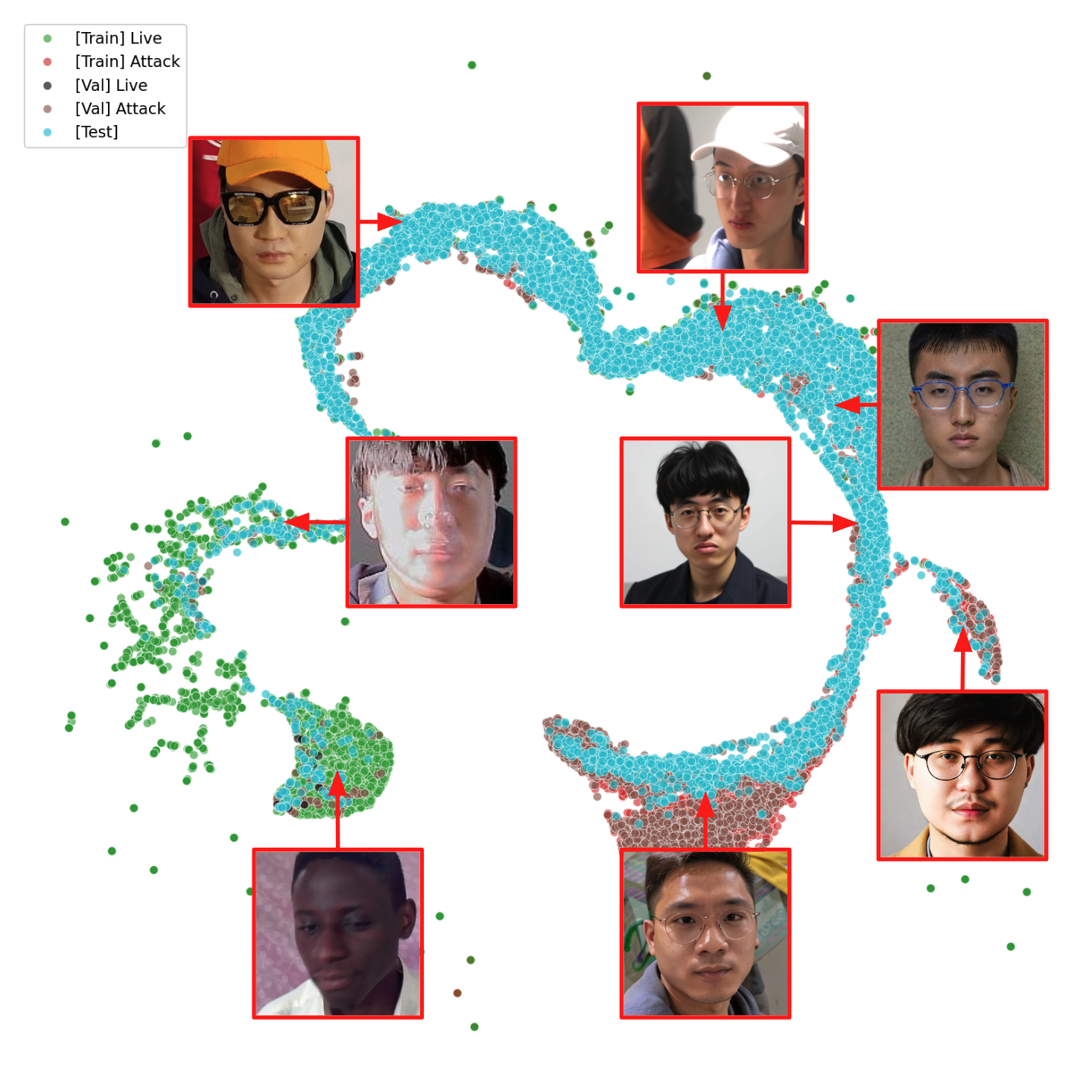}
    \caption{UMAP projection grouped by dataset split.}
    \label{fig:umap-test}
  \end{subfigure}
  \caption{UMAP projections used in the ablation study. Example images are shown to illustrate the types of attacks present in each cluster.}
  \label{fig:umap-combined}
\end{figure*}
  


\textbf{Data filtering threshold}. To determine the optimal cosine similarity threshold $\tau_{\text{sim}}$ for pairing attack samples with their most similar live counterparts, we conducted a grid search over threshold values from 0.84 to 0.91. Lower thresholds admit more attack--live pairs into the training set, increasing dataset size but also allowing a higher proportion of loosely matched identities. Therefore, higher thresholds ensure stronger identity consistency between pairs, but overly restrictive filtering risks removing diverse training examples. As shown in Tab.~\ref{tab:threshold}, the threshold $\tau_{\text{sim}} = 0.90$ yielded the lowest ACER (1.08\%) and the highest accuracy (97.85\%) on the validation set. This indicates that stricter matching leads to more consistent identity pairs, enabling the model to focus on liveness-related cues. Based on these results, we adopted $\tau_{\text{sim}} = 0.90$ for our experiments.  

\begin{table}[t]
\centering
\small
\renewcommand{\arraystretch}{1.05}
\begin{tabularx}{\columnwidth}{c c c c c}
\toprule
\textbf{Threshold} & \textbf{Dataset Size} & \textbf{ACER\,$\downarrow$} & \textbf{Acc\,$\uparrow$} & \textbf{AUC\,$\uparrow$} \\
\midrule
0.84 & 10339 & 0.1426 & 0.8686 & 0.9550 \\
0.85 & 10131 & 0.0661 & 0.9446 & 0.9749 \\
0.86 & 9887  & 0.0621 & 0.9527 & 0.9850 \\
0.87 & 9637  & 0.0785 & 0.9199 & 0.9673 \\
0.88 & 9362  & 0.0789 & 0.9192 & 0.9706 \\
0.89 & 9036  & 0.0769 & 0.9231 & 0.9747 \\
0.90 & 8669  & \textbf{0.0108} & \textbf{0.9785} & 0.9903 \\
0.91 & 8207  & 0.0594 & 0.9581 & \textbf{0.9908} \\
\bottomrule
\end{tabularx}
\caption{Effect of cosine similarity threshold on validation performance in the paired-sampling filtering stage.}
\label{tab:threshold}
\end{table}

\begin{figure}[h]
    \centering
    \includegraphics[width=\linewidth]{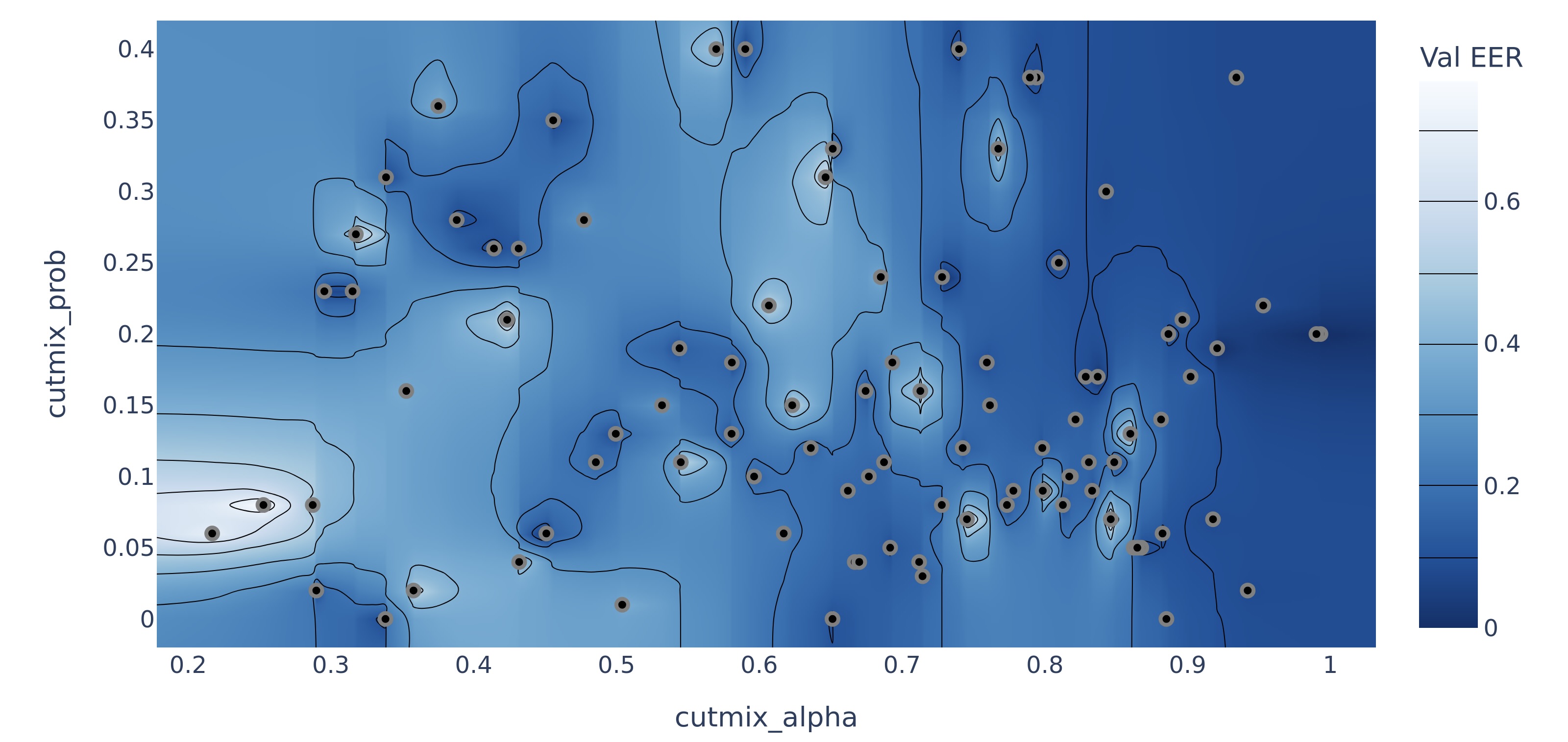}
    \vspace{1mm}
    \includegraphics[width=\linewidth]{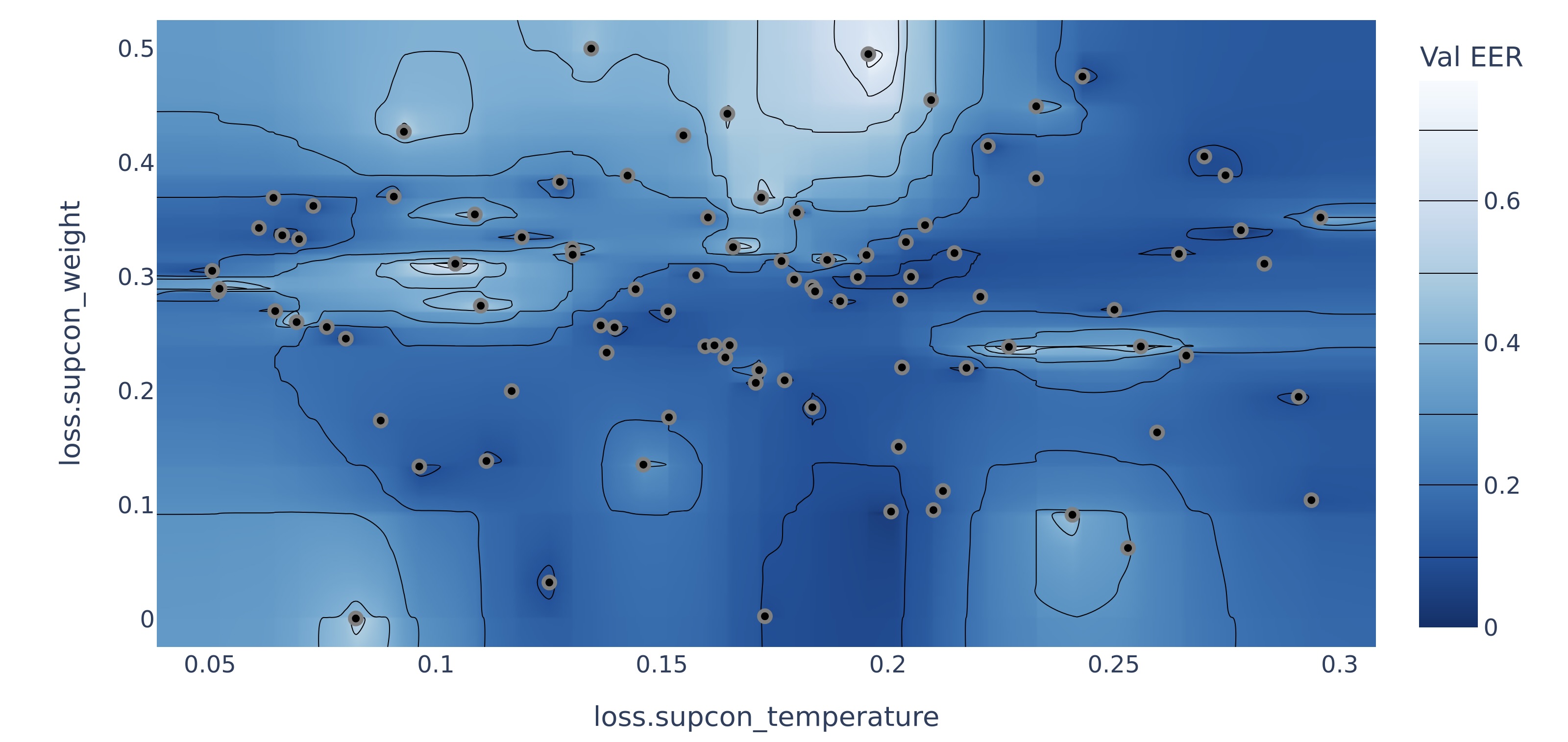}
    \caption{Impact of CutMix parameters (\textbf{top}) and supervised contrastive loss parameters (\textbf{bottom}) on validation EER. 
    Darker regions indicate lower EER values. 
    The CutMix plot varies $\alpha$ and probability $p$, while the SupCon plot varies loss weight $\lambda_{\mathrm{supcon}}$ and temperature $T$.}
    \label{fig:hyperparam-impact}
\end{figure}

\paragraph{Hyperparameter Sensitivity Analysis.}
To assess the influence of key augmentation and contrastive learning parameters on validation performance, we conducted a grid search over the CutMix parameters $\alpha$ and $p$, as well as the supervised contrastive loss weight $\lambda_{\mathrm{supcon}}$ and temperature~$T$. 
Fig.~\ref{fig:hyperparam-impact} visualizes the validation EER as a function of these hyperparameters. 

The top plot reveals that low to moderate CutMix probabilities $p \approx 0.15$--$0.3$ combined with $\alpha \approx 0.6$--$1.0$ tend to minimize EER, suggesting that excessive CutMix usage or overly aggressive blending may obscure liveness cues. 
On the bottom, the optimal supervised contrastive configuration is found near $\lambda_{\mathrm{supcon}} \approx 0.2$--$0.3$ and $T \approx 0.15$--$0.2$, balancing feature compactness for genuine samples with sufficient separation from attack embeddings. 
These results informed the final hyperparameter settings used in our challenge submission.

\textbf{UMAP analysis}. To better understand the distribution of live and attack samples, we conducted a UMAP analysis on the learned feature representations.

Fig.~\ref{fig:umap-val} presents a UMAP projection of various spoof types. Each point corresponds to a sample in the dataset, color-coded according to the type of manipulation. 

We observe a clustering of live samples, which are separated from the majority of attack types. Additionally, certain spoof categories tend to form localized clusters, indicating that the model learns discriminative patterns not only between live and fake samples, but also among different kinds of attacks. This suggests that some attack types share similar visual or statistical characteristics in the learned feature space. Example images are overlaid to provide visual context for these clusters.

Fig.~\ref{fig:umap-test} shows a UMAP projection plot where the samples are grouped according to dataset split and label. We observe significant overlap between live samples from the training set and visually obvious attacks from the validation and test sets. This suggests that despite apparent differences from a human perception perspective, the learned representation sometimes struggles to draw clear boundaries, probably due to the limited variability in training data or the complexity of generalizing across different distributions. They also highlight which spoof types may be more challenging for generalization in the proposed approach.

\begin{figure}[ht]
  \centering
  \setlength{\tabcolsep}{2pt}
  \begin{tabular}{ccc}
    \includegraphics[width=0.32\linewidth]{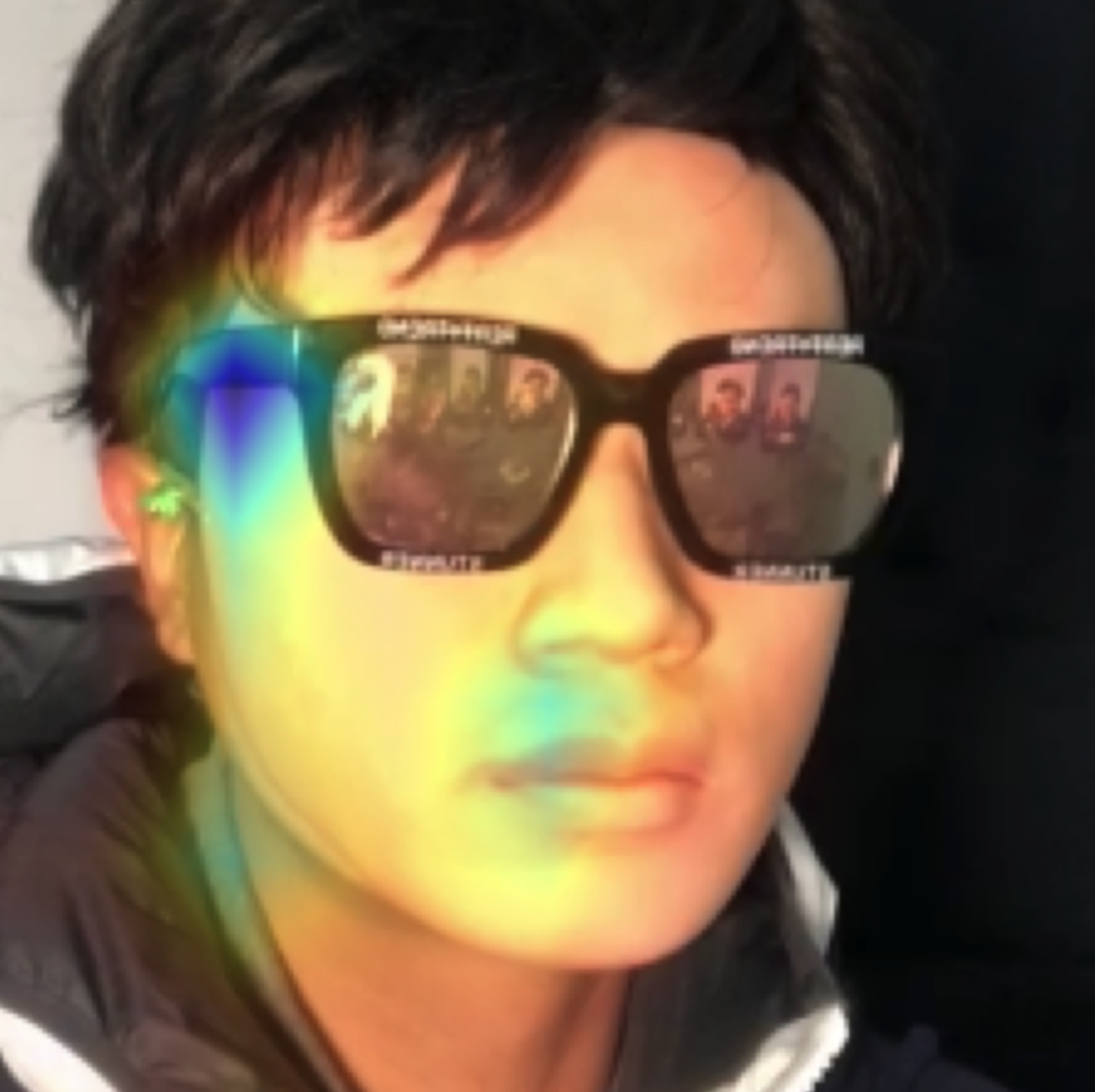} &
    \includegraphics[width=0.32\linewidth]{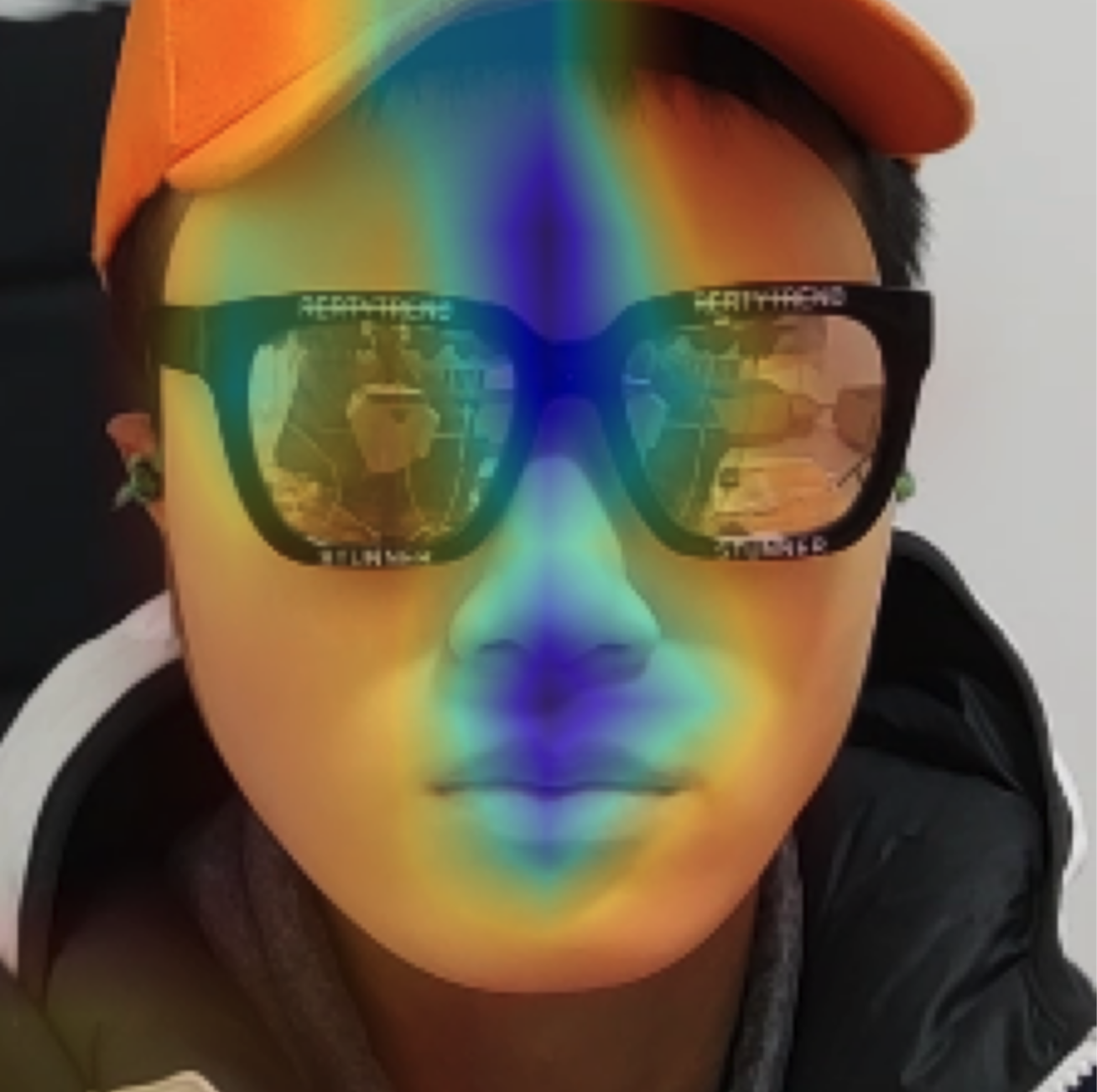} &
    \includegraphics[width=0.32\linewidth]{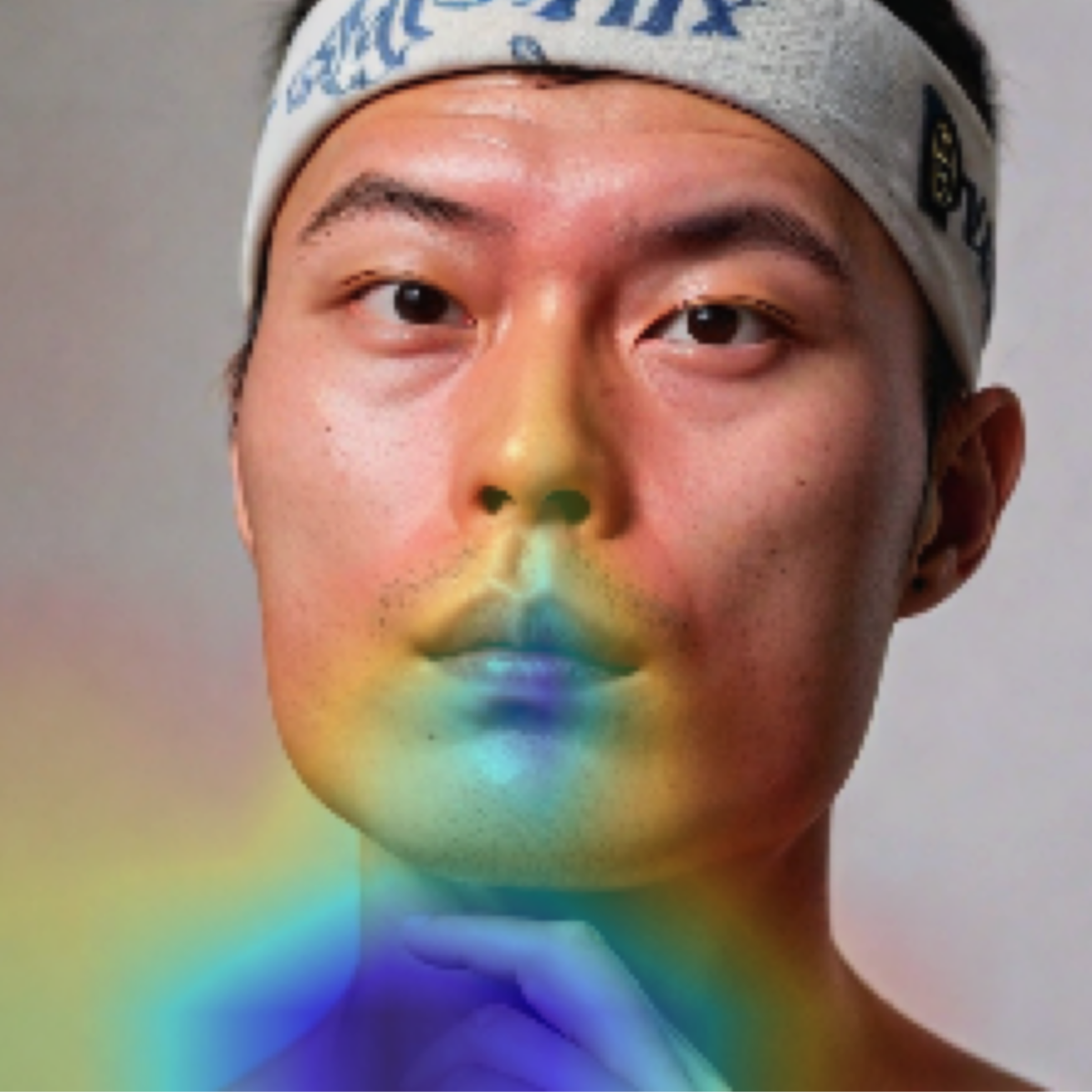} \\
    (a) & (b) & (c)
  \end{tabular}
  \caption{Model attention on attacks.
  The train detector focuses on human-visible cues: mask edge discontinuities (1), specular highlights (b), and skin texture anomalies (c)}
  \label{fig:ablation_heatmaps}
\end{figure}

\textbf{Model cues}. On UniAttackData our model mostly attends to cues that are also obvious to a human observer.
As shown in Fig.~\ref{fig:ablation_heatmaps}, (b) and (c) highlight specular reflections and texture inconsistencies on the glasses/skin, while (a) focuses on the mask boundary and local edge discontinuities.

\section{Conclusion}

In this paper, we presented a paired-sampled contrastive framework for face attack detection that jointly addresses both physical presentation attacks and digital forgeries within a single model. Our contrastive parallel training strategy leverages automatically paired live and spoof images, and trained via a combined focal and contrastive loss in order to learn robust liveness representations. We implemented this approach using a lightweight ConvNeXt-v2-Tiny backbone with the CutMix technique, and achieved an ACER of 2.10\% in "The 6th Face Anti-Spoofing: Unified Physical-Digital Attacks Detection@ICCV 2025" challenge, placing it among the top-performing methods.

Extensive experiments show that our method achieves high detection accuracy ACER of 2.10\% on the challenge dataset, while maintaining low computational overhead (4.46 GFLOPs) suitable for real‐time deployment. Moreover, our analysis highlights the benefit of contrastive learning and asymmetric augmentation in improving generalization to unseen attack types and domains.

We believe that a unified detection system, informed by both presentation and digital attack cues, will be critical for securing next‐generation face recognition systems against evolving threats.

{
    \small
    \bibliographystyle{ieeenat_fullname}
    \bibliography{main}
    
}

\end{document}